\documentclass[runningheads]{llncs}
\usepackage{graphicx}
\usepackage{booktabs,tabularx}
\usepackage{multirow}
\usepackage{amssymb}
\usepackage{color}

\usepackage{makecell}
\usepackage{ulem}
\usepackage{marvosym}

\usepackage[table]{xcolor}
\begin{document}
\title{Semantic Graph Representation Learning for Handwritten Mathematical Expression Recognition}
\titlerunning{Semantic Graph Representation Learning for HMER}

\author{Zhuang Liu\inst{1}\orcidID{0009-0007-1105-7052} \and
Ye Yuan\inst{1}\orcidID{0000-0002-2822-1564} \and
Zhilong Ji\inst{1}\orcidID{0000-0002-8799-3409} \and
Jinfeng Bai\inst{1}\orcidID{0000-0001-8940-480X} \and
Xiang Bai\inst{2}\textsuperscript{(\Letter)}\orcidID{0000-0002-3449-5940}}

\authorrunning{Z. Liu et al.}
\institute{
Tomorrow Advancing Life \\
\email{\{liuzhuang7, jizhilong\}@tal.com}, 
\email{yuanye\_phy@hotmail.com}, 
\email{jfbai.bit@gmail.com}
\and Huazhong University of Science and Technology \\
\email{xbai@hust.edu.cn}
}

%
%
%
\maketitle

\begin{abstract}

Handwritten mathematical expression recognition (HMER) has attracted extensive attention recently. However, current methods cannot explicitly study the interactions between different symbols, which may fail when faced similar symbols. To alleviate this issue, we propose a simple but efficient method to enhance semantic interaction learning (SIL). Specifically, we firstly construct a semantic graph based on the statistical symbol co-occurrence probabilities. Then we design a semantic aware module (SAM), which projects the visual and classification feature into semantic space. The cosine distance between different projected vectors indicates the correlation between symbols. And jointly optimizing HMER and SIL can explicitly enhances the model's understanding of symbol relationships. In addition, SAM can be easily plugged into existing attention-based models for HMER and consistently bring improvement. Extensive experiments on public benchmark datasets demonstrate that our proposed module can effectively enhance the recognition performance. Our method achieves better recognition performance than prior arts on both CROHME and HME100K datasets. 

\end{abstract}


%
\section{Introduction}

Handwritten Mathematical Expression Recognition (HMER) is an important OCR task, which can be widely applied in question parsing and answer sheet correction. In recent years, with the rapid development of deep learning technology, scene text recognition approaches have achieved great progress \cite{shi2016end,shi2018aster,yue2020robustscanner,fang2021read}. However, due to the ambiguities brought by crabbed handwriting and the complicated structures of handwritten mathematical expressions, HMER is still a challenging task.

Built upon the recent progress in sequence-to-sequence learning and neural networks \cite{sutskever2014sequence,chung2014empirical,huang2017densely}, some studies have addressed HMER with end-to-end trained encoder-decoder models and showed significant improvement in performance. Nevertheless, the encoder-decoder framework do not fully explore the correlation between different symbols in the mathematical expression, which may be struggling when facing similar handwritten symbols or crabbed handwritings.

\begin{figure*}[t]
	\centering
	\includegraphics[width=0.8\textwidth]{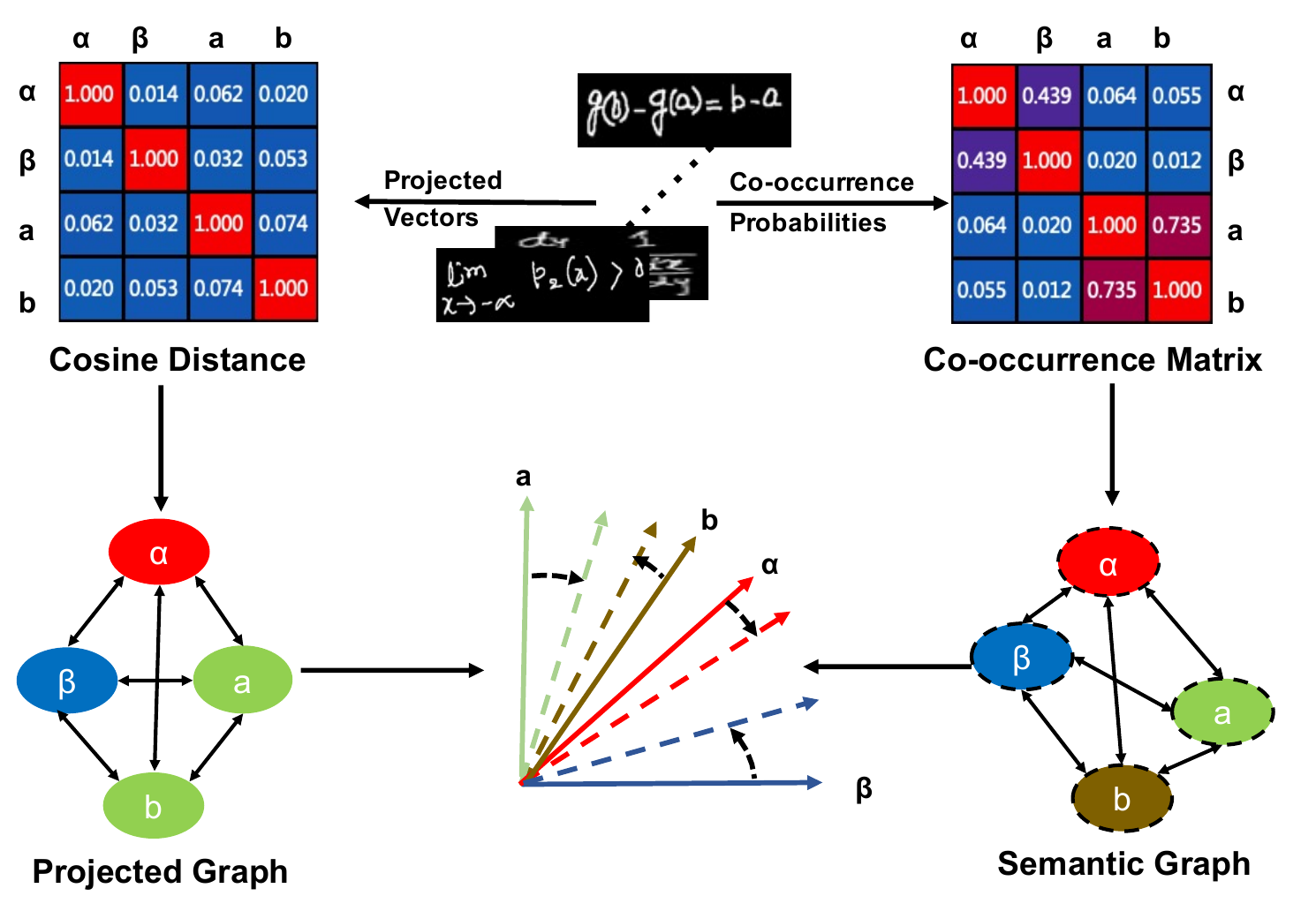}
	\caption{Illustration of our method. The different colored graph nodes and arrows indicate different symbols.}
	\label{fig:image}
\end{figure*}

To address above issues, we argue that an effective HMER model should be improved from the following two aspects: (1) capturing semantic dependencies among different symbols in the mathematical expression; (2) integrating more semantic information to locate the regions of interest.

In this paper, we propose an simple but efficient method to improve the robustness of the model, which incorporate the learning of semantic relations among different symbols into the end-to-end training (Fig. \ref{fig:image}). Firstly, we built a semantic graph rely on statistical co-occurrence probabilities, which can explicitly exhibit the dependencies among different symbols. Secondly, we propose a semantic aware module, which takes the visual and classification features as input and maps them into the semantic space. The cosine distance between different projected vectors suggests the correlation of symbols. Optimizing the distance to close to the corresponding graph value make the network capture the relationships between different symbols. Therefore, the search for regions of interest and the learning of symbols semantic dependencies are enhanced, which further improved the performance of the model.

The major contributions of this paper are briefly summarized as follows:
\begin{itemize}
    \item To the best of our knowledge, we are the first to use co-occurrence to represent the relationship between symbols in mathematical expression and verify the effectiveness of enhancing semantic representation learning.
    \item We propose a semantic aware method that jointly optimizes the symbol relations learning and HMER, which can consistently improve the performance of the model for HMER.
    \item Our proposed semantic aware module can be easily plugged into attention based models for HMER and no extra computation during the inference stage.
\end{itemize}

To be specific about the performance, we adopt DWAP \cite{zhang2017watch} as the baseline network. With the help of SAM, SAM-DWAP outperforms DWAP by 2.2\%, 2.8\% and 4.2\% on CROHME 2014, 2016 and 2019, respectively. Moreover, with adopting the latest SOTA method CAN \cite{li2022counting} as the baseline network, our method achieves new SOTA results (58.0\% on CROHME 2014, 56.7\% on CROHME 2016, 58.0\% on CROHME 2019). This indicates that our method can be generalized to various existing encoder-decoder models for HMER and boost their performance.

\section{Related Work}

HMER is a fundamental OCR task, which has attracted research interests in the past several decades. 
In this section, we briefly introduce previous related works on HMER.

Traditional methods on HMER could be mainly separated into two steps: a symbol segmentation/recognition step and a grammar guided structure analysis step. In the first step,
several classic classification techniques were studied, such as HMM \cite{winkler1996hmm,kosmala1999line,hu2011hmm,alvaro2014recognition}, Elastic Matching \cite{chan1998elastic,vuong2010towards}, Support Vector Machines \cite{keshari2007hybrid}, etc. 
In the second step, formal grammars were designed to model the 2D and syntactic structures of expression. 
Lavirotte \emph{et al.} \cite{lavirotte1998mathematical} proposed to use graph grammar to recognize mathematical expression. 
Chan \emph{et al.} \cite{chan2001error} incorporated correction mechanism into parser based on definite clause grammar (DCG). 
Yamamoto \emph{et al.} \cite{yamamoto2006line} modeled handwritten mathematical expressions with a stochastic context-free grammar and solved the recognition problem by using the CYK algorithm. 
In contrast to those traditional methods, our model incorporates grammatical structure and automatically learned encoder-decoder, therefore preventing from designing cumbersome rules.

Recently, deep learning techniques rapidly boosted the performance of HEMR. 
The mainstream framework was encoder-decoder networks \cite{deng2017image,zhang2018multi,zhang2017watch,zhang2018track,zhao2021handwritten,wang2019multi,truong2020improvement,nguyen2021temporal,le2020recognizing,zhang2018track,le2019pattern,wu2021graph}. 
Deng \emph{et al.} \cite{deng2017image} firstly proposed an encoder-decoder framework to convert image to \LaTeX\ markup. 
A coarse-to-fine attention layer was used to reduce the attention complexity in their work. 
Zhang \emph{et al.} \cite{zhang2017watch} presented an encoder-decoder model, named WAP (Watch, Attend and Parse).
In their model, the encoder is a FCN and a coverage vector is appended to the attention model. 
Wu \emph{et al.} \cite{wu2018image,wu2020handwritten} focused on the pair-wise adversarial learning strategy to improve the recognition accuracy.
To alleviate the challenge of lack of data, Le \emph{et al.} \cite{le2019pattern} and Li \emph{et al.} \cite{li2020improving} employed distortion, decomposition and scale augmentation techniques, which achieved significant performance promotion. 
Le \cite{le2020recognizing} proposed a dual loss attention model, which contains a new context match loss. Context matching loss is adapted to constrain the intra-class distance and enhance the discriminative power of model.
Lately, Zhang \emph{et al.} \cite{zhang2020tree} devised a tree-based decoder to parse mathematical expression. 
At each step, a parent and child node pair was generated and the relation between parent node and child node reflects the structure type. Yuan \emph{et al.} \cite{yuan2022syntax} firstly incorporate syntax information into the encoder-decoder, which achieved higer recognition accuracy while taking into account speed. Li \emph{et al.}\cite{li2022counting} design a weakly-supervised counting module and jointly optimizes HMER task and symbol counting task. With the help of integrated global information, it puts in a impressive performance.

\section{Methodology}

\begin{figure*}[t]
	\centering
	\includegraphics[width=1\textwidth]{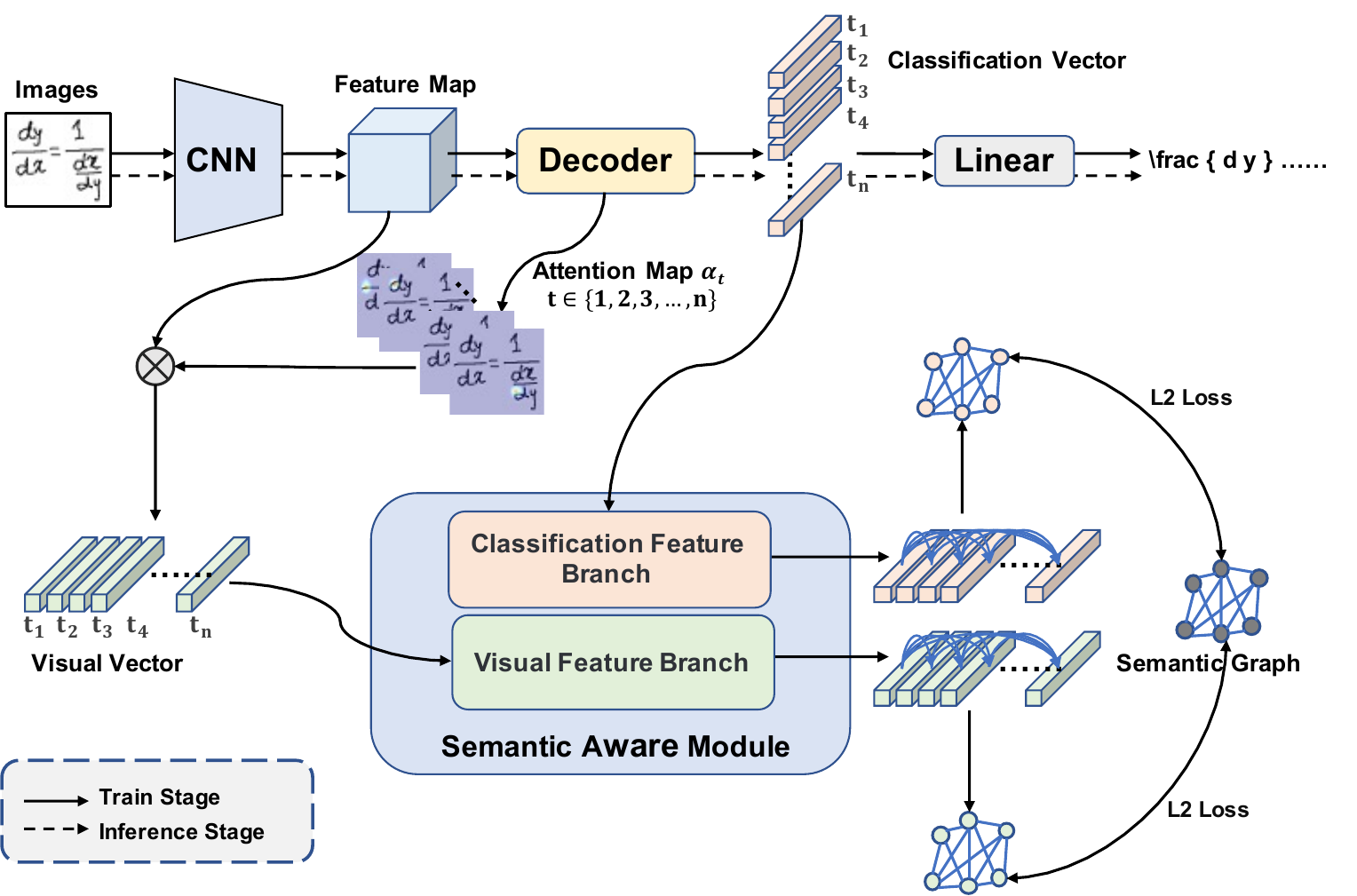}
	\caption{The architecture of the proposed SAM-DWAP, which consists of a CNN, a decoder and a semantic aware module.}
	\label{fig:model}
\end{figure*}

The overall framework of our approach is shown in Fig. \ref{fig:model}. The pipeline includes several parts: densely connected convolutional network (DenseNet) \cite{huang2017densely} is applied as encoder to extract the features. The DenseNet takes a grayscale image X of size $H \times W \times 1$, where $H$ and $W$ are image height and image width, respectively, and returns a 2D feature map $\mathcal{F} \in \mathbb{R}^{H ^ {'} \times W ^ {'}\times 684}$, where $H/H^{'} = W/W^{'} = 16$. The decoder uses the feature map and gradually predicts the \LaTeX\ markup. The \textbf{S}emantic \textbf{A}ware \textbf{M}odule (SAM) comprises two branches with similar structure (visual branch and classification branch), which employ the visual and classification features, respectively. Visual and classification features are projected to semantic space to obtain projected visual and classification vectors, respectively. The cosine distance between projected vectors from different time steps indicates how related they are.


\subsection{Semantic Graph}
Capturing global context information has been proven to be an effective way to improve the robustness of recognition \cite{yu2020towards,qiao2020seed}. However, compared with words, the use of symbols in the mathematical expressions is relatively more casual. How to express the relationships among different symbols in the mathematical expressions is an open issue to be solved. Our intuition is that the magnitude of values in the co-occurrence graph reflects the relationship between different symbols, much like how different characters in text have different collocations. Making the distances close to the probabilities is aimed at enhancing the model's learning of the linguistic information in formulas.

Semantic graph is defined as $G = (S, E)$, where $S = \{s_{1}, s_{2}, ..., s_{N} \}$ represents the set of symbol nodes and $E$ represents the edges, which suggest the dependence between any two symbols. 
The correlation matrix $R = \{r_{i,j}\}_{i,j=1}^{N}$ of graph $G$ contains non-negative weights associated with each edge. The correlation matrix is a conditional probability matrix and the $r_{ij}$ is set as $P(s_i/s_j)$, where $P$ is calculated through training set. However, $R$ is an asymmetric matrix, namely $r_{ij} \neq r_{ji}$. In order to facilitate the calculation, we turn the asymmetric matrix into a symmetric matrix following:
\begin{equation}
    R^{'} = \frac{1}{2} (R + R^T).
\end{equation}

\begin{figure*}
	\centering
 	\includegraphics[width=0.95\textwidth]{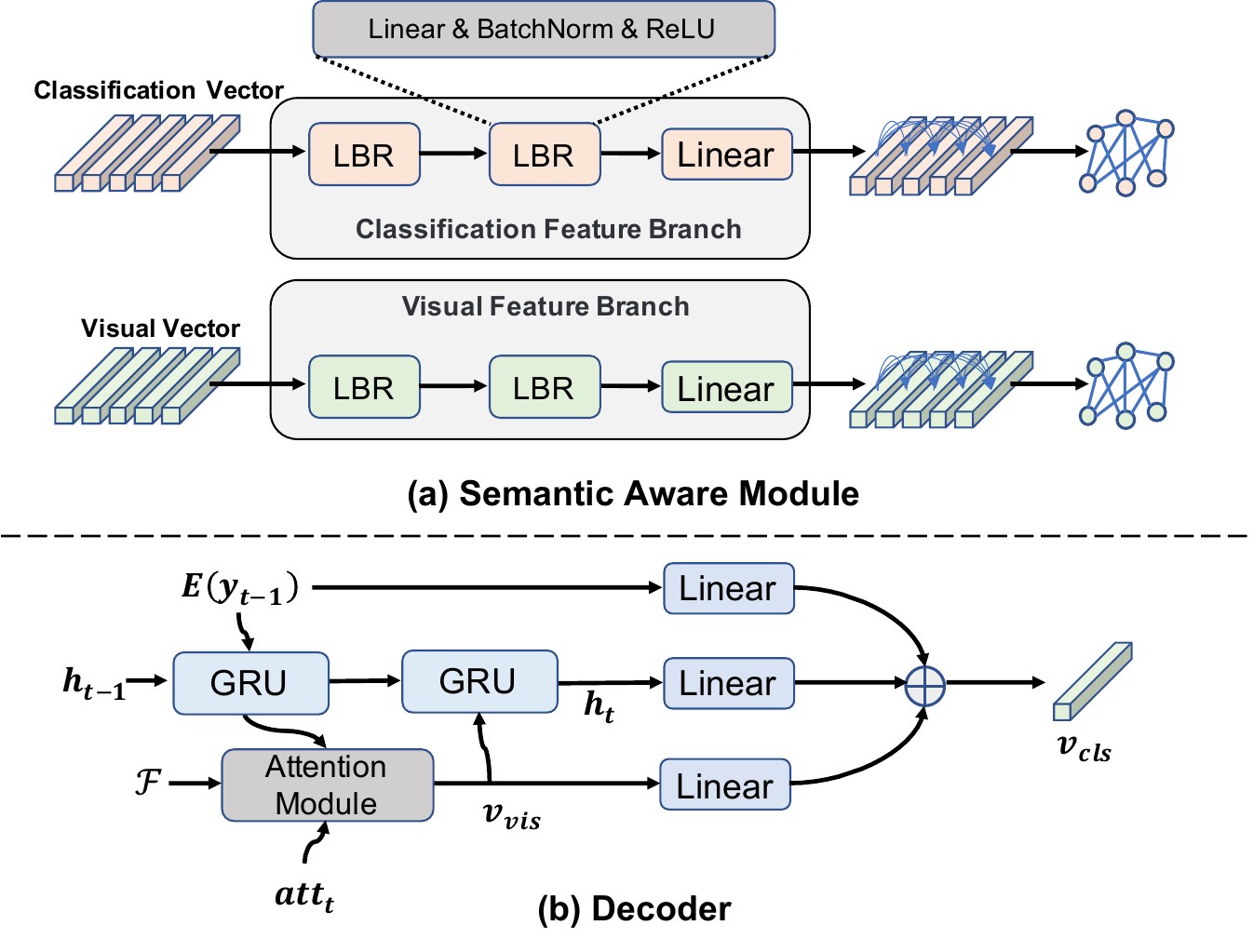}
	\caption{The architecture of (a) semantic aware module (SAM) and (b) decoder. 
 } \label{fig:sim}
\end{figure*}
\subsection{Semantic Aware Module} \label{sub:SIM}

In this section, we present the detail of the proposed semantic aware module (SAM). As shown in Fig. \ref{fig:sim} (a), SAM contains two branches, namely visual feature branch and classification feature branch. Each branch comprises two ``LBR'' block followed by a linear layer. A ``LBR'' block is built by stacking \textbf{L}inear layer, \textbf{B}atch Normalization and \textbf{R}eLU activation. We apply SAM to project the visual vectors ($v_{vis}$) and classification vector ($v_{cls}$) to semantic space to get projected visual vectors ($v_{vis}^{'}$) and projected classification vectors ($v_{cls}^{'}$):
\begin{equation}
    v_{vis}^{'} = W_3^{vis}(\sigma(\epsilon(W_2^{vis}(\sigma(\epsilon(W_1^{vis}v_{vis} 
 + b_1^{vis}))) + b_2^{vis} ))) + b_3^{vis}
\end{equation}
\begin{equation}
    v_{cls}^{'} = W_3^{cls}(\sigma(\epsilon(W_2^{cls}(\sigma(\epsilon(W_1^{cls}v_{cls} 
 + b_1^{cls}))) + b_2^{cls} ))) + b_3^{cls}
\end{equation}
where $\sigma$ is the ReLU activation and $\epsilon$ refers to Batch Normalization. $W_3^{vis}$, $W_2^{vis}$, $W_1^{vis}$, $W_3^{cls}$, $W_2^{cls}$ and $W_1^{cls}$ are learnable parameters.

Our goal is to optimize the projected visual vectors ($v_{vis}^{'}$) and projected classification vectors ($v_{cls}^{'}$). Such that cos($v_{i}^{'}$, $v_{j}^{'}$) is close to $R_{ij}$ for all i, j, where cos($v_{i}^{'}$, $v_{j}^{'}$) denotes the cosine similarity between $v_{i}^{'}$ and $v_{j}^{'}$:

\begin{equation}
    cos(v_{i}^{'}, v_{j}^{'}) = \frac{v_{i}^{'T}v_{j}}{||v_{i}^{'T}||\ ||v_{j}^{'T}||}
\end{equation}

\subsection{Decoder} \label{sub:decoder}

Fig. \ref{fig:sim} (b) shows the structure of decoder. The decoder mainly contains two Gated Recurrent Units (GRU) cells and an attention module. The first GRU takes the symbol embedding ($E(y_{t-1})$) and historical state ($h_{t-1}$) predicted in the last step as input and output a new hidden state vector $h_{t}^{'}$:

\begin{equation}
    h_t^{'} = GRU(E(y_{t-1}), h_{t-1})
\end{equation}
Then the attention module calculates the attentional weights $\alpha_t$ through its attention mechanism:
\begin{equation}
    e_{t} = W_{\omega}(tanh(W_{h}^{'}h_t^{'} + W_{f}\mathcal{F} + W_{\alpha}att_t)) 
\end{equation}
\begin{equation}
    \alpha_{t} = exp(e_{t}) / \sum exp(e_{t})
\end{equation}
where $W_{\omega}$, $W_{h}^{'}$, $W_{f}$ and $W_{\alpha}$ are trainable parameters. $\mathcal{F}$ represents the feature map and $att_t$ refers to coverage attention \cite{zhang2017watch}, which equals the sum of all past attention
probabilities:
\begin{equation}
    att_t = \sum \limits _{i} \alpha_{i}, \ \ \ i \in [0, t-1]
\end{equation}
The $\alpha_{t}$ and $\mathcal{F}$ are multiplied to obtain visual features vectors $v_{vis}$:
\begin{equation}
    v_{vis} = \alpha_{t} \otimes \mathcal{F}
\end{equation}
The second GRU takes the $v_{vis}$ and $h_t^{'}$ as input and returns the hidden state $h_{t}$:
\begin{equation}
    h_t = GRU(v_{vis}, h_t^{'})
\end{equation}
Then we aggregate $E(y_{t-1})$, $v_{vis}$ and $h_t$ to obtain the classification feature vectors and symbol probabilities:
\begin{equation}
    v_{cls} = W_{e}E(y_{t-1}) + W_{h}h_t + W_{v}v_{vis} 
\end{equation}
\begin{equation}
    p_{symbol} = softmax(W_{s}v_{cls})
\end{equation}
where $W_{e}$, $W_{h}$, $W_{v}$ and $W_{s}$ are trainable parameters.

\subsection{Loss Function} \label{sub:loss}
The overall function consists of three parts and is defined as follows:
\begin{equation}\label{formula:loss}
    \mathcal{L} = \mathcal{L}_{symbol} + \mathcal{L}_{vis} + \mathcal{L}_{cls}
\end{equation}
where $\mathcal{L}_{symbol}$ is cross entropy classification loss of the predicted probability $p_{symbol}$ with respect to its ground-truth. $\mathcal{L}_{vis}$ and $\mathcal{L}_{cls}$ are L2 regression loss defined as follows:
\begin{equation}
\mathcal{L}_{vis} = \sum_{i}^{n} \sum_{j}^{n} (cos(v_{vis,i}, v_{vis, j}),  - R_{i,j}) ^ {2}
\end{equation}
\begin{equation}
\mathcal{L}_{cls} = \sum_{i}^{n} \sum_{j}^{n} (cos(v_{cls,i}, v_{cls, j}),  - R_{i,j}) ^ {2}
\end{equation}

\section{Experiments}

We conduct experiments on three CROHME and HME100K benchmark datasets and compare the performance with previous state-of-the-art methods. In this section, we firstly specify the datasets, implementation details and evaluation protocol in Section \ref{dataset}, \ref{implementation details} and \ref{protocol}, respectively. Then, in Section \ref{compare with sota} we evaluate our method on public datasets and compare it with other state-of-the-art methods. In Section \ref{ab studies}, we exhibit the ablation studies and finally, in Section \ref{case studies} we show few cases and discuss the effectiveness of our method.

\subsection{Datasets} \label{dataset}
\subsubsection{CROHME Dataset.} CROHME dataset is from the competition on recognition of online handwritten mathematical expression, which is the most widely used public dataset. Images in CROHME dataset are synthesized from the handwritten stroke trajectory information in the InkML files. Therefore, the image background from CROHME dataset is clean (Fig. \ref{fig:HME100K} (a)). The CROHME training set number is 8,836, while the test set contains 986, 1147 and 1199 images respectively due to different release years.

\subsubsection{HME100K Dataset.} 
\begin{figure*}[t]
\begin{center}
  \includegraphics[width=0.96\linewidth]{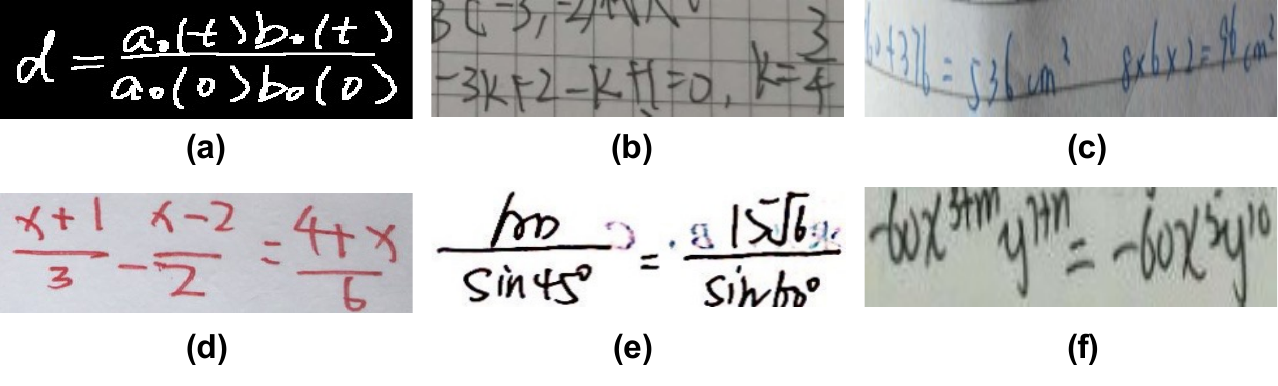}
\end{center}
\vspace{-2ex}
\caption{Sample images from (a) CROHME dataset and (b-f) HME100K dataset.}
\label{fig:HME100K}
\end{figure*}

HME100K dataset is a real scene dataset and consequently, HME100K dataset are varied in color, blur, complicated background, twist (Fig. \ref{fig:HME100K} (b-f)). HME100K dataset contains 74,502 images for training and 24,607 images for testing. The data size of HME100K dataset is ten times larger than CROHME dataset. The number of math symbols included in the HME100K dataset is 245, which is two times larger than that of CROHME dataset.

\subsection{Implementation Details} \label{implementation details}
The proposed methods is implemented in PyTorch. A single Nvidia Tesla V100 with 32GB RAM is used to conduct experiment. The batch size is set at 8.
Both the hidden state sizes of the two GRUs and dimension of word embedding are set at 256. The Adadelta optimizer \cite{zeiler2012adadelta} is used during the training process, in which $\rho$ is set at 0.95 and $\epsilon$ is set at $10^{-6}$. The learning rate starts from 0 and monotonously increases to 1 at the end of the first epoch. After that the learning rate decays to 0 following the cosine schedules \cite{zhang2019bag}. For CROHME dataset, the total training epoch is set to 240 and for HME100K dataset, the training epoch is set to 40.

\begin{table}[t]
\setlength\tabcolsep{3pt}   
\small
\caption{Expression Recognition Rate (ExpRate) performance of SAM-DWAP and SAM-CAN and other
state-of-the-art methods on CROHME 2014, CROHME 2016 and CROHME 2019 test set. SAM-DWAP and SAM-CAN indicate adopting DWAP and CAN as the backbone, respectively. All results are reported as a percentage (\%). }
\centering
\begin{tabular}{c|cccc}
\hline
Method & \makecell[c]{CROHME\\2014} & \makecell[c]{CROHME\\2016} & \makecell[c]{CROHME\\2019} & \makecell[c]{HME100K} \\
\hline\hline
UPV~\cite{DBLP:conf/icfhr/MouchereVZG14} & 37.22 & - & - & -\\
TOKYO~\cite{DBLP:conf/icfhr/MouchereVZG16} & - & 43.94 & - & - \\
PAL~\cite{DBLP:conf/pkdd/WuYZZL18} & 39.66 & - & - & - \\
WAP~\cite{DBLP:journals/pr/ZhangDZLHHWD17} & 46.55 & 44.55 & - & - \\
PAL-v2~\cite{DBLP:journals/ijcv/WuYZZL20} & 48.88 & 49.61 & - & -\\
TAP~\cite{DBLP:journals/tmm/ZhangDD19} & 48.47 & 44.81 & - & -\\
DLA~\cite{DBLP:conf/cvpr/Le20} & 49.85 & 47.34 & - & -\\
DWAP~\cite{DBLP:conf/icpr/ZhangDD18} & 50.10 & 47.50 & - & 61.85 \\
DWAP-TD~\cite{DBLP:conf/icml/ZhangDYSW020} & 49.10 & 48.50 & 51.40 & 62.60\\
DWAP-MSA~\cite{DBLP:conf/icpr/ZhangDD18} & 52.80 & 50.10 & 47.70 & - \\
WS-WAP~\cite{DBLP:conf/icfhr/TruongNPN20} & 53.65 & 51.96 & - & -\\
MAN~\cite{DBLP:conf/icdar/WangDZW19} & 54.05 & 50.56 & - & - \\
BTTR~\cite{DBLP:conf/icdar/ZhaoGYPDZ21} & 53.96 & 52.31 & 52.96 & 64.10 \\
SAN~\cite{anderson1967syntax} & 56.20 & 53.60 & 53.50 & 67.10 \\
ABM~\cite{abm} & 56.85 & 52.92 & 53.96 &  65.93 \\
CAN~\cite{li2022counting} & 57.00 & 56.06 & 54.88 & 67.31 \\
\hline
\textbf{SAM-DWAP (ours)} & 56.80 & 55.62 & 56.21 & 68.08\\
\textbf{SAM-CAN (ours)} & \textbf{58.01} & \textbf{56.67} &  \textbf{57.96} & \textbf{68.81}\\
\hline
\end{tabular}
\label{table:sota}
\end{table}

\subsection{Evaluation Protocol} \label{protocol}
\textbf{Recognition Protocol.} We employ expression recognition rate (ExpRate) to evaluate the performance of different approaches. The definition of ExpRate is the percentage of predicted mathematical expressions that exactly match the ground truth.

\subsection{Comparison with State-of-the-Art} \label{compare with sota}
\subsubsection{Results on the CROHME Datasets.}

Tab. \ref{table:sota} summaries the performance of our method and previous methods on the CROHME dataset. Since most of the previous work does not use data augmentation, we mainly discuss the results without data augmentation.

As shown in tab. \ref{table:sota}, using DWAP \cite{zhang2018multi} as the backbone, SAM-DWAP achieves competitive results to the last SOTA method CAN \cite{li2022counting} on CROHME 2014 and CROHME 2016. On CROHME 2019 dataset, our method ourperforms CAN by 1.33 \%.

To further verify our proposed SAM is compatible with other models and can consistently bring performance improvements. We integrate SAM into CAN to construct SAM-CAN. As shown in tab. \ref{table:sota}, SAM-CAN achieves the best performance on all CROHME test set and outperforms CAN by 1.21 \%, 0.61 \% and 3.08 \%, respectively. This result clearly demonstrates the effectiveness of our proposed module.



\subsubsection{Results on the HME100K Dataset.}

\begin{table}[t]
    \centering
    \caption{Performance of SAM-DWAP and SAM-CAN versus DWAP, DWAP-TD, BTTR, ABM, SAN and CAN on the HME100K dataset on Easy (E.), Moderate (M.) and Hard (H.) HME100K test subsets. Our models achieve the best performance on the HME100K dataset.}
    \label{subset performance}
    \begin{tabular}{|m{.3\columnwidth}|m{.12\columnwidth}<{\centering}|m{.14\columnwidth}<{\centering}|m{.12\columnwidth}<{\centering}|m{.12\columnwidth}<{\centering}|}
    \hline
    \textbf{HME100K}& \textbf{Easy}& \textbf{Moderate}& \textbf{Hard}& \textbf{Total} \\
    \hline
    \textbf{Image size}&       7721&   10450&   6436& 24607\\
    \hline\hline
    DWAP\cite{zhang2018multi}&    75.1&  62.2& 45.4& 61.9\\
    DWAP-TD\cite{zhang2020tree}& 76.2& 63.2& 45.4& 62.6\\
    BTTR \cite{zhao2021handwritten}& 77.6 & 65.3 & 46.0 & 64.1\\
    ABM~\cite{abm} & - & - & - & 65.3 \\
    SAN~\cite{yuan2022syntax}& 79.2 &  67.6 & 51.5 & 67.1 \\
    CAN~\cite{li2022counting} & - & - & - & 67.3 \\
    \hline
    \textbf{SAM-DWAP(ours)} & 79.3 & 68.4 & 54.0 & 68.1 \\
    \textbf{SAM-CAN(ours)} & \textbf{79.8} & \textbf{69.8} & \textbf{54.0} & \textbf{68.8}\\
    \hline
    \end{tabular}
    \label{100k results}
\end{table}
\begin{table*}[t]
\small
\caption{Ablation Studies on CROHME dataset. DWAP$^\dagger$ and CAN$^\dagger$ are our reproduced results. The effect of recognition performance with regard to the two components: visual feature branch and classification feature branch.}
\centering
\begin{tabular}{|m{.30\columnwidth}|m{.12\columnwidth}<{\centering}|m{.12\columnwidth}<{\centering}|m{.12\columnwidth}<{\centering}|}
\hline
\multirow{2}*{Method}& \multicolumn{3}{c|}{CROHME}\\
\cline{2-4}
 & 2014 &  2016 & 2019 \\
\hline\hline
DWAP$^\dagger$ & 54.6 & 52.8 & 52.0 \\
Vis-DWAP & 55.8 & 54.8 & 54.1 \\
Cls-DWAP & 55.6 & 55.2 & 54.7 \\
\textbf{SAM-DWAP} & \textbf{56.8} & \textbf{55.6} & \textbf{56.2} \\
\hline\hline
CAN$^\dagger$ & 57.1 & 55.3 & 54.9\\
Vis-CAN & 57.5 & 56.3 & 56.6 \\
Cls-CAN & 57.4 & 56.5 & 55.8 \\
\textbf{SAM-CAN} & \textbf{58.0} & \textbf{56.6} & \textbf{57.9} \\
\hline
\end{tabular}
\vspace{+1ex}
\label{tab:ablation-study}
\end{table*}

As shown in tab. \ref{table:sota} and \ref{100k results}, we compare our prosposed method with DWAP \cite{zhang2018multi}, DWAP-TD \cite{zhang2020tree}, BTTR \cite{zhao2021handwritten}, ABM \cite{abm}, SAN \cite{yuan2022syntax} and CAN \cite{li2022counting} on HME100K dataset. It is clear to notice that SAM-DWAP and SAM-CAN achieves the best performance. Specifically, as shown in table \ref{100k results}, SAM-DWAP and SAM-CAN outperform SAN by 0.1 \% and 0.6 \% on easy subset, respectively. However, as the difficulty of the test subset increases, the
leading margin of our method increases to 2.5 \% and 2.5 \% on the hard subset. This further proves the effectiveness of the proposed SAM.


\begin{figure}[t]
	\centering
	\includegraphics[width=1\textwidth]{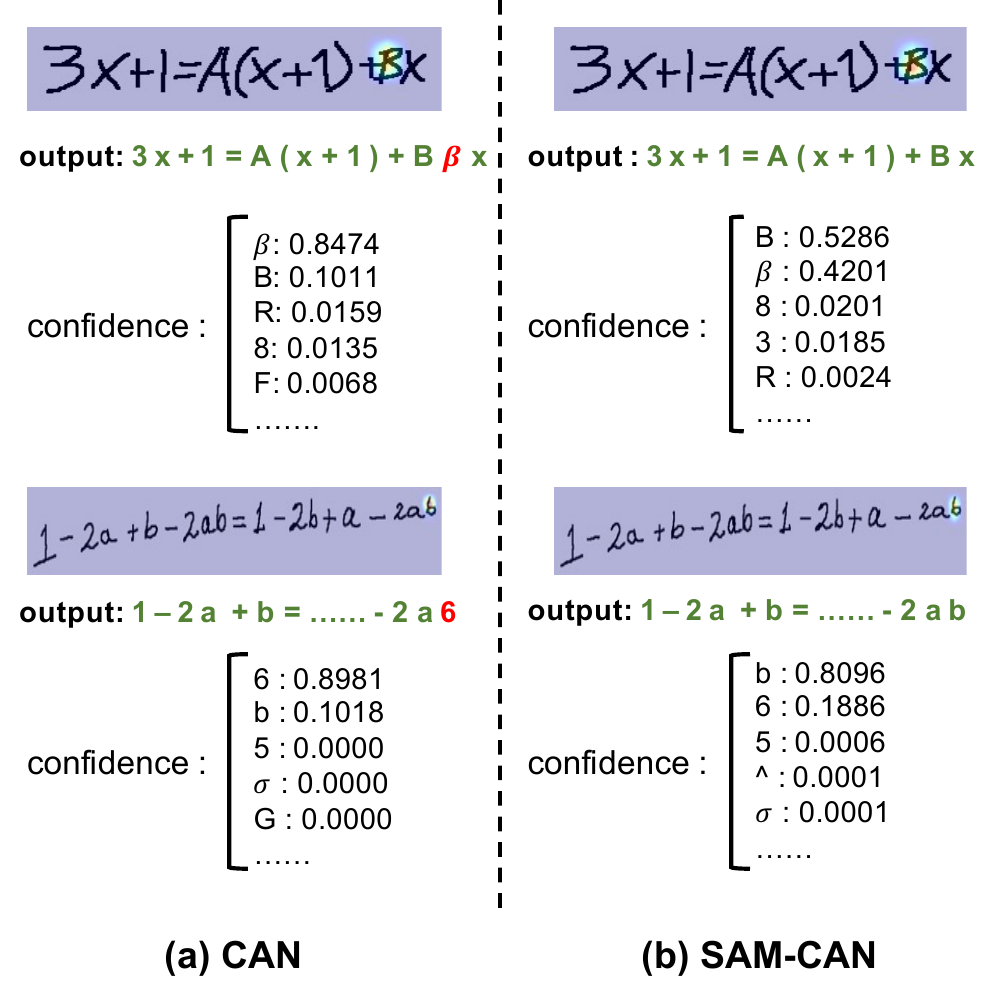}
	\caption{Examples of (a) CAN and (b) SAM-CAN. Symbol in red color are mispredictions}
	\label{fig:case}
\end{figure}

\subsection{Ablation Study}\label{sub:AStudy} \label{ab studies}

In this subsection, we evaluate the effectiveness of visual feature branch and classification feature branch. SAM-DWAP and SAM-CAN are the default models. Vis-DWAP and Vis-CAN have a visual feature branch but not a classification feature branch. Cls-DWAP and Cls-CAN have a classification feature branch but not a visual feature branch. DWAP$^\dagger$ and CAN $^\dagger$ are our reproduced results. The results are summarized in tab. \ref{tab:ablation-study}.

\subsubsection{Impact of Visual Feature Branch.} Tab. \ref{ab studies} shows adopting visual feature
branch to DWAP improves the recognition performance ExpRate by 1.2 \% on CROHME 2014, 2.0 \% on CROHME 2016 and 2.1 \% on CROHME 2019. Inserting visual feature branch into CAN also can enhance the performance by 0.4 \% on CROHME 2014, 1.0 \% on CROHME 2016 and 1.7 \% on CROHME 2019. Hence integrating visual feature branch can effectively improve the performance.

\subsubsection{Impact of Classification Feature Branch.} Tab. \ref{ab studies} shows adopting classification feature branch to DWAP improves the recognition performance ExpRate by 1.0 \% on CROHME 2014, 2.4 \% on CROHME 2016 and 2.7 \% on CROHME 2019. Inserting visual feature branch into CAN also can enhance the performance by 0.3 \% on CROHME 2014, 1.2 \% on CROHME 2016 and 0.9 \% on CROHME 2019. Hence integrating classification feature branch can effectively improve the performance.

\subsection{Case Study} \label{case studies}

In this section, we show two examples to illustrate the effect of using SAM. As shown in Fig. \ref{fig:case} (a), although CAN correctly focuses on the region of interest, it misidentifies the symbol ``B'' as symbol ``$\beta$'' and misidentifies the symbol ``b'' as symbol ``6''. In contrast, the regions of interest of the SAM-CAN are similar to those of CAN, but SAM-CAN correctly predicts symbol ``B'' and ``b''. The confidences of symbol ``B'' and ``b'' also increase from 10.1 \% to 52.9 \% and increase from 10.2\% to 81.0\%, respectively. This phenomenon indicates that adopting SAM can improve the robustness of recognition especially the recognition performance of similar symbols.



\section{Conclusion}

This paper has presented a simple and efficient method for handwritten mathematical expression recognition by incorporate semantic graph representation learning into end-to-end training. To our best knowledge, the proposed method is the first to learn the correlation between different symbols through symbol co-occurrence probabilities. Experiments on the CROHME dataset and HME100K dataset have validated the effectiveness and efficiency of our method.

\section{Acknowledgement}
This work was supported by National Key R\&D Program of China, under Grant No. 2020AAA0104500 and National Science Fund for Distinguished Young Scholars of China (Grant No.62225603).

\bibliographystyle{splncs04}
\bibliography{mybibliography}





\end{document}